%% file: main.tex
\documentclass{article}
\usepackage{ijcai19}

\usepackage{times}
\usepackage{soul}
\usepackage{url}
\usepackage[hidelinks]{hyperref}
\usepackage[utf8]{inputenc}
\usepackage[small]{caption}
\usepackage{graphicx}
\usepackage{subcaption}
\usepackage{amsmath}
\usepackage{amsfonts}
\usepackage{bbm}
\usepackage{booktabs}
\usepackage{algorithm}
\usepackage{algorithmic}
\title{Successor Options: An Option Discovery \\ Framework for Reinforcement Learning}

\author{
Manan Tomar$^1$
\and
Rahul Ramesh$^1$\And
Balaraman Ravindran$^{1}$
\affiliations
$^1$Indian Institute of Technology Madras
\emails
rahul13ramesh@gmail.com, manan.tomar@gmail.com, ravi@cse.iitm.ac.in
}

\begin{document}

\maketitle

\input{sections/AbstractIntroPrelim.tex}

\input{sections/method.tex}

\input{sections/exp.tex}
\input{sections/related.tex}

\newpage
\bibliographystyle{named}
\bibliography{ijcai19}

\end{document}

%% file: sections/AbstractIntroPrelim.tex
\begin{abstract}
The options framework in reinforcement learning models the notion of a skill or a temporally extended sequence of actions. The discovery of a reusable set of skills has typically entailed building options, that navigate to \textit{bottleneck} states. This work adopts a complementary approach, where we attempt to discover options that navigate to \textit{landmark states}. These states are prototypical representatives of well-connected regions and can hence access the associated region with relative ease. In this work, we propose \textit{Successor Options}, which leverages Successor Representations to build a model of the state space. The intra-option policies are learnt using a novel pseudo-reward and the model scales to high-dimensional spaces easily. Additionally, we also propose an \textit{Incremental Successor Options} model that iterates between constructing Successor Representations and building options, which is useful when robust Successor Representations cannot be built solely from primitive actions. We demonstrate the efficacy of our approach on a collection of grid-worlds, and on the high-dimensional robotic control environment of Fetch.
\end{abstract}

\section{Introduction}
Reinforcement Learning (RL) \cite{sutton1998reinforcement} has garnered significant attention recently due to its success in challenging high-dimensional tasks \cite{mnih2015humanlevel,lillicrap2015continuous,schulman2015trust}. Deep Learning has had a major role in the achievements of RL by enabling generalization across a large number of states using powerful function approximators. Deep learning must however be complemented by efficient exploration in order to discover solutions with reasonable sample complexities. \textit{Hierarchical Reinforcement Learning} (HRL) is one potential strategy that mitigates the curse of dimensionality by operating on abstract state and action spaces. Recent work \cite{vezhnevets2017feudal,kulkarni2016hierarchical,bacon2016optioncritic} has attempted to use a hierarchy of controllers, operating in different time-scales, in order to search large state spaces rapidly. 

\begin{figure}[t]
    \centering
    \includegraphics[scale=0.32]{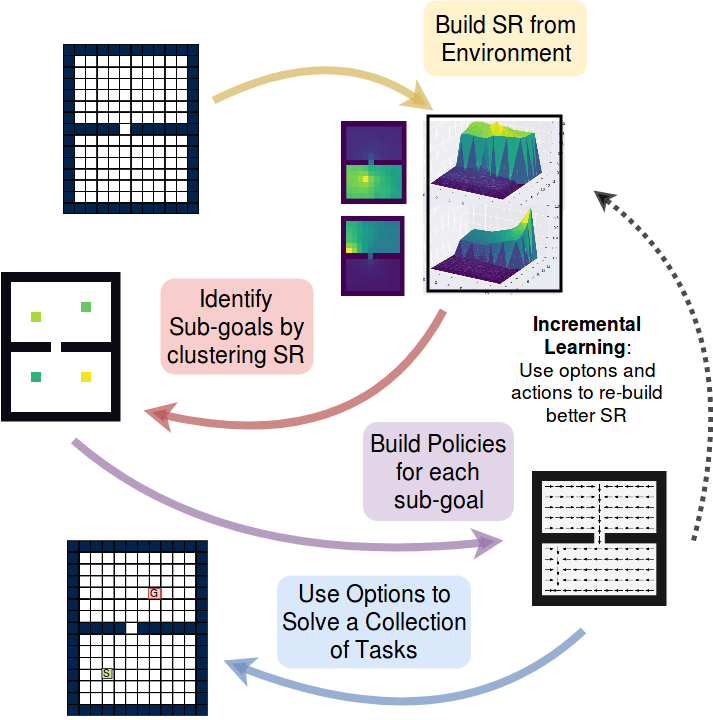}
    \caption{An overview of Successor Options Framework}
    \label{fig:intro}
\end{figure}

The \textit{options} framework \cite{sutton1999mdps} is an example of a hierarchical approach that models temporally extended actions or \textit{skills}. Discovering ``good" options can potentially allow for exploring the state space efficiently and transferring to various similar tasks. However, the discovery of reusable options is a meticulous task and has not been effectively addressed. While there are a number of approaches to this problem, a large fraction of literature revolves around discovering options that navigate to \textit{bottleneck states} \cite{mcgovern2001automatic,csimcsek2004using,simsek2005identifying,menache2002q}. This work adopts a paradigm, that fundamentally differs from the idea of identifying bottleneck states as sub-goals for options. Instead, we attempt to discover landmark or \textit{prototypical states} of well-connected regions. We empirically validate that navigating to landmark states leaves the agent well situated, to navigate the associated well-connected region consequently. Building on this intuition, we propose \textit{Successor Options}, a sub-goal discovery, and intra-option policy learning framework. 

Our method does not construct a graph of the state space explicitly but instead leverages \textit{Successor Representations} (SR) \cite{dayan1993improving}, to learn the sub-goals and the associated intra-option policies. The SR inherently captures the temporal structure between states and thus forms a reasonable proxy for the actual graph. Moreover, Successor Representations have been extended to work with function approximators \cite{kulkarni2016deep,barreto2017successor}, allowing us to implicitly form the graphical structure of any high-dimensional state space with neural networks.

SRs attempt to assign similar representations to states with similar future states. Formally, the SR of a state $s$ is a vector representing the expected discounted visitation counts of all states in the future, starting from state $s$. SR varies with the policy since the expected visitation counts depend on the policy being executed, to estimate these counts. Since nearby states are expected to have similar successors, their Successor Representations are expected to be similar in nature. Hence, states in a well-connected region in state space will have similar SRs (for example, states in a single room in a grid-world will have very similar SRs). Building on this intuition, one would like to identify a set of sub-goals for which the corresponding SRs are dissimilar to one another. 

Successor Options proceeds as follows. The first step involves constructing the SRs of all states. The sub-goals are then identified by clustering a large sample (or all) of the SR vectors and assigning the cluster centers as the various sub-goals. 
The cluster centers translate to a set of sub-goals that have vastly different successor states, meaning different sub-goals provide access to a different region in state-space. Once the sub-goals are identified, a novel \textit{pseudo-reward} is used to build options that navigate to each of these sub-goals. This process relies solely on primitive actions to navigate the state space when estimating the SRs. However, in large state spaces, full exploration through primitive actions might not be possible. To mitigate the same, we propose the \textit{Incremental Successor Options} algorithm. This method works in an iterative fashion where each iteration involves an option discovery step and an SR learning step. 

Besides the improved accessibility to any given state in the state space, Successor Options offer a number of other advantages over existing option discovery methods. While an intermediate clustering step segments the algorithm into distinct stages (non-differentiability introduced), the step is critical in many aspects. Firstly, the number of options $k$ is specified beforehand which allows the model to adapt by finding the $k$ most suited sub-goals. Hence, the algorithm does not require pruning redundant options from a very large set, unlike other works \cite{mcgovern2001automatic,csimcsek2009skill,machado2017laplacian}. Furthermore, the discovered options are reward agnostic and are hence transferable across multiple tasks. 
The principal contributions of this paper are as follows : \textbf{(i)} An automatic option discovery mechanism revolving around identifying landmark states, \textbf{(ii)} A novel pseudo reward for learning the intra-option policies that extends to function approximators \textbf{(iii)} An incremental approach that alternates between exploration and option construction to navigate the state space in tasks with a fixed horizon setup where primitive actions fail to explore fully. 

\section{Preliminaries}

Reinforcement Learning deals with sequential decision making tasks and considers the interaction of an agent with an environment. It is traditionally modeled by a Markov Decision Process (MDP) \cite{puterman1994markov}, defined by the tuple $\mathcal{\langle S, A, P, \rho_{0}}, r, \gamma \rangle$, where $\mathcal{S}$ defines the set of states, $\mathcal{A}$ the set of actions, $\mathcal{P  : S \times A} \rightarrow \mathcal{P}(S)$ the transition function, $\mathcal{\rho_{0}}$ the probability distribution over initial states, $r\mathcal{ : S \times S ' \times A \rightarrow R}$ the reward function and $\gamma$ the discount factor. In the context of optimal control, the objective is to learn a policy that maximizes the discounted return $R_{t} = \sum_{i=t}^{T} \ \gamma^{(i-t)} \ r(s_{i}, s_{i+1}, a_{i})$, where $r(s_{i}, s_{i+1}, a_{i}) \ $ is the reward function.

{\bf Q-learning:} Q-learning \cite{watkins1992q} attempts to estimate the optimal action-value function $Q^*(s,a)$. It exploits the Bellman optimality equation, the repeated application of which leads to convergence to $Q^*(s,a)$. The optimal value function can be used to behave optimally by selecting action $a$ in every state such that $a \in \text{argmax}_{a'} Q(s,a')$

\begin{align} 
Q(s_t, a_t) \leftarrow Q(s_t, a_t) &+ \alpha \Big[ r_{t+1} +  \nonumber\\
   &\gamma \max \limits_{a'} Q(s_{t+1}, a') - Q(s_t, a_t) \Big ]
\end{align}

\cite{mnih2015humanlevel} introduce Deep Q-learning, that extends Q-learning to high dimensional spaces by using a neural network to model $Q_{\theta}(s,a)$

%\subsection{Options and Semi-Markov Decision Processes}

{\bf Options and Semi-Markov Decision Processes:} Options \cite{sutton1999mdps} provide a framework to model temporally extended actions. Formally, an option is defined using the 3-state tuple : $\mathcal{\langle I, \beta, \pi \rangle}$, where $\mathcal{I} \subseteq \mathcal{S}$ is the initiation set, $\beta : \mathcal{S} \rightarrow [0, 1]$ the termination probabilities for each state and $\pi : \mathcal{S} \rightarrow P(\mathcal{A})$ the intra-option policy. This work assumes that the intra-option policies satisfy the Markov assumption. 

{\bf Successor Representation:} The Successor Representation (SR) \cite{dayan1993improving} represents a state $s$ in terms of its successors. The SR for $s$ is defined as a vector of size $|\mathcal{S}|$ with the $i^{th}$ index equal to the discounted future occupancy for state $s_i$ given the agent starts from $s$. Since the SR captures the visitation of successor states, it is directly dependent on the policy $\pi$ and the transition dynamics $p(s_{t+1} | s_{t}, a_{t})$. More concretely, the SR can be written as follows:

\begin{figure}
    \centering
    \includegraphics[scale=0.25]{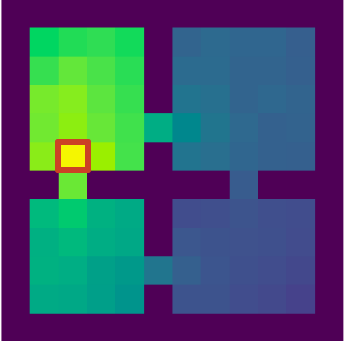}
    \hspace{0.2cm} 
    \includegraphics[scale=0.21]{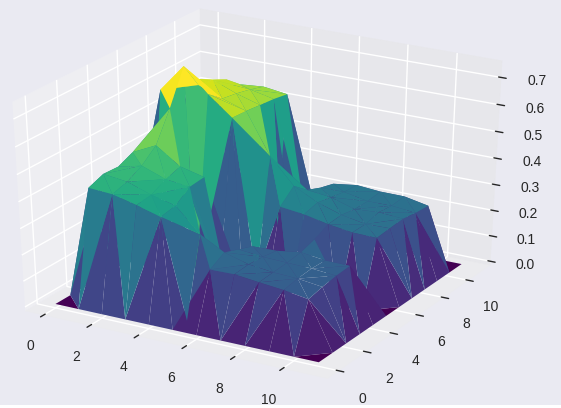}
    \caption{Successor Representation of state $s$ (marked in red) under the uniform random policy. The SR vector is projected onto 2D grid for the purpose of visualization.}
    \label{fig:srviz}
\end{figure}

\begin{equation} 
\psi_{\pi}(s, s^{'}) = \mathbb{E}_{s^{'} \sim P, a \sim \pi} \ \left[\sum_{t=0}^{\infty} \ \gamma^{t} \mathbb{I}(s_{t} = s^{'}) \ | \ s_{0} = s \right]
\end{equation}
where, $\mathbb{I(.)}$ is 1 if its argument is true, else 0 (indicator function).  The SR can be learnt in a temporal difference (TD) like fashion by writing it in terms of the SR of the next state.

\begin{equation} \label{eq:1}
\hat{\psi}(s_t, :) \leftarrow \hat{\psi}(s_t, :)  \ +  \ \alpha \left[\mathbbm{1}_{s_t} + \gamma [\hat{\psi}(s_{t+1}, :)]- \hat{\psi}(s_t, :)\right] 
\end{equation}

Equation \ref{eq:1} is for state samples $s_t, s_{t+1}$, where $\hat{\psi}$ is the estimate of SR being learnt and $\mathbbm{1}_{s_t}$ is a one-hot vector with all zeros except a 1, at the $s_t^{th}$ position. Successor Representations can be naturally extended to the deep setting \cite{kulkarni2016deep,barreto2017successor} as follows (note $\phi(s)$ is k-dimensional feature representation of $s$, and $\theta$ is the set of parameters) :

\begin{equation} \label{eq:2}
\psi_{\pi}(s_t; \theta) = \mathbb{E} \ \left[ \phi(s_t) + \gamma \ \psi_{\pi}(s_{t+1} ; \theta)  \right]
\end{equation}

%% file: sections/method.tex
\section{Proposed Method}

Successor Options (SR-Options) adopts an approach that attempts to discover options that navigate to states that are representatives of well-connected regions. The method holds a number of advantages which include \textbf{(i)} A robust sub-goal identification step that uses clustering to obtain a set of options, with no two options being identical. \textbf{(ii)} Learning useful options without an extrinsic reward, but through latent learning \textbf{(iii)} Using an incremental approach to work in scenarios where primitive actions are unable to facilitate the option discovery process.

\subsection{Successor options}

\textbf{Sub-goal discovery}:  In learning Successor Options, the first step involves learning the SR. The policy used to learn the SR ($\pi_{SR}$) determines a prior over state space. As a result, the discovered sub-goals will lie in those states which are more likely to be visited under $\pi_{SR}$. Since we do not have any such preference in our experiments, we stick to the uniform random policy for $\pi_{SR}$ in this work.

This is followed by clustering states, based on the learnt SR (we utilize K-means++ \cite{arthur2007k} for this purpose). Since the SR captures temporally close-by states efficiently, the generated clusters are spread across the state space, with each cluster assigned to a set of states that are densely connected. We wish to learn options that navigate to the cluster centers which act as landmark states. Since the cluster center may not correspond to the SR of any state, we select the sub-goal to be that state whose SR has the largest cosine similarity with the cluster center.

\textbf{Latent Learning:} The pseudo reward defined in Equation \ref{eq:eprew} is used to learn the intra-option policies.

\begin{equation} \label{eq:eprew}
    r_{\psi_s}(s_t, a_t, s_{t+1}) = \psi_{g_k}(s_{t+1}) - \psi_{g_k}(s_t)
\end{equation}

For Equation \ref{eq:eprew}, an agent transitions from state $s_t$ to state $s_{t+1}$ under action $a_t$. $\psi_{g_k}(s_{t})$ is the $s_t$\textsuperscript{th} component of the SR vector of sub-goal $g_k$ ($g_k$ is $k^{th}$ cluster center). $\psi_{g_k}(s_{t})$ can also be understood as the discounted visitation count of state $s_t$, starting from state $g_k$. Hence, the reward is proportional to the change in the discounted visitation counts of the states involved in the transition. Why this reward? The reward drives the agent to states which have highest values of $\psi_{g_k}(s_{t})$, meaning they are led to states that have the highest visitation count when starting from state $g_k$. Hence the agent is driven to \textit{landmark states} by the pseudo-reward and can be understood as a hill-climbing task on the SR (see Figure \ref{fig:srviz}). The option policy terminates when the agent reaches the state with the highest value of $\psi_{g_k}(s)$ which occurs when Value function $V_{g_k}^*(s) \leq 0$ (this condition is used to decide option termination). 

Hence, every sub-goal has a corresponding pseudo-reward, that navigates the agent to that sub-goal before terminating. Furthermore, this reward is not hand-crafted and is dense in nature (see Figure \ref{fig:srviz}), which leads to faster learning. Note that an approximately developed SR is often a sufficient signal to learn the optimal policy for the option. Formally, the initiation set for the options is the set of all states $\mathcal{S}$, the option deterministically terminates ($\beta=1$) at states with $V(s) \leq 0$ (all other states have $\beta=0$) and the option policy is dictated by the reward function in Equation \ref{eq:eprew}.

\textbf{Solving Tasks}: The learnt options can be used under an SMDP framework to solve tasks that differ in their reward structure. One can use SMDP-Q-learning with intra-option value function updates \cite{sutton1999mdps} for faster learning, since the learnt options are Markovian in nature.

\subsection{Incremental Successor Options}

SR-options relies on primitive actions to build the SR of all states. Finite horizon environments are good examples of scenarios where the SR cannot be learnt with a uniform random policy, which would consequently lead to poor options. Hence, we propose an incremental approach where we discover intermediate options, which facilitate the SR learning process. Such an approach would be critical in long-horizon tasks, where the exploration can be facilitated using reward agnostic options.

The algorithm (See Algorithm \ref{alg:algorithm}) starts by building the SR from primitive actions. This is followed by an option discovery step from the current SR matrix. In the next iteration, the option and actions are used in tandem to construct a more robust SR. Since we are interested in the SR of the uniformly random policy, the SR is not updated when executing an option, but only when executing primitive actions (update refers to Equation \ref{eq:1}). The constructed intermediate options can be used in any manner but one would ideally want to sample actions more frequently than options since options navigate to specific sub-goals and sampling them frequently would hence limit you to certain states. After the SR is rebuilt, a new set of options are formed with this SR and the old set is discarded. The newly formed options are used in the next iteration and the process is repeated. Finally, when the SRs are sufficiently built, one can use SR-options, with the final SR matrix obtained from the incremental exploration procedure.

% Algorithm 
%%%%%%%%%%%%% 
\begin{algorithm}[ht]
\caption{Incremental SR-Options}
\textbf{Input} : \textit{N}: Number of iterations to form SR \\
\textbf{Input} : \textit{k}: Number of options 
\label{alg:algorithm}
\begin{algorithmic}[1] %[1] enables line numbers
    \STATE A $\leftarrow$ $ \{ a_0, a_1, \cdots, a_n \}$ where $a_i \in \mathcal{A}$ 
    \STATE O $\leftarrow \{ \}$, $\Psi \leftarrow \text{Zero Matrix}$
    \FOR{ i = 1, ...., N }
        \STATE $\Psi \leftarrow$ UpdateSR(O+A, $\Psi$)
        \STATE $\text{CandS} \leftarrow$ GetCandidateStates($\Psi, SR_{min}, SR_{max}$)
        \STATE $\text{sub-goals} \leftarrow$ ClusterSR($\text{CandS}, \Psi$)
        \STATE O $\leftarrow \{ \}$
        \FOR{ g in $\text{sub-goals}$}
            \STATE o $\leftarrow$ LearnOption($\Psi(g)$)
            \STATE store o in O
        \ENDFOR
    \ENDFOR
    \STATE O $\leftarrow$ SR-Options($\Psi$, k)
    \STATE \textbf{return} O
\end{algorithmic}
\end{algorithm}

How are the options obtained in the incremental setup? Ideally, one would like to discover options that drive the agent towards unvisited parts of state space. While visitation count would be one such ideal metric, we make use of the L1-norm of the SR-vector as a proxy for the visitation count \cite{machado2018count}.  Hence, only states with low L1-norm SRs, participate in the clustering. As shown in Algorithm \ref{alg:algorithm}, the clustering stage uses a set of candidate sub-goals which are a fraction of the set of reached states. Formally, a state $s$ is a candidate sub-goal if $SR_{min} <|\psi(s)|_1 < SR_{max}$, where $SR_{min}$ and $SR_{max}$ are hyper-parameters that decide the range of L1-norms of the selected states. Such a condition ensures that all candidate sub-goal states have an SR that is neither fully developed nor extremely sparse or underdeveloped, thus providing a pseudo reward which is easy to learn over. 

\subsection{Deep Successor Options}

\begin{figure}[ht]
    \centering
    \includegraphics[width=0.45\textwidth]{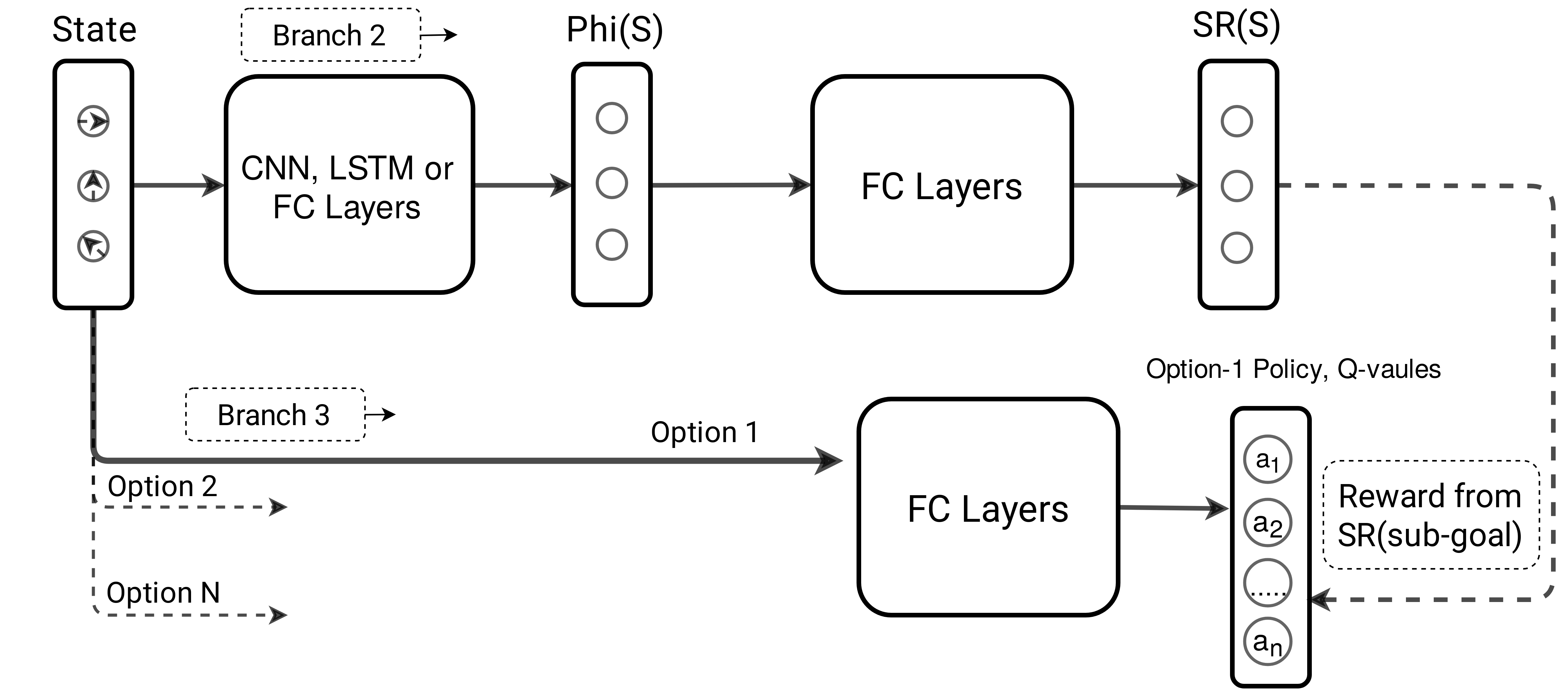}
    \caption{Neural Net for Deep Successor Options. The Architecture is trained in two stages. The first step involves learning the SF. The second stage learns optimal policies/Q-value functions from the pseudo rewards obtained from the SF.}
    \label{fig:net}
\end{figure}

Deep Successor Options (Figure \ref{fig:net}) extends SR-options to the function approximation setting. \cite{kulkarni2016deep,barreto2017successor} propose Successor Features (SF), a model for generating the SR using neural networks. Deep Successor options is extended to continuous action spaces by learning through three branches. These branches are the reward prediction error (branch 1), the TD error for learning Successor Features (branch 2) and the Option policy heads (branch 3). The first two branches usually share the same base representation $\phi(s)$. However, reward prediction is required only when one is interested in computing the Q-values. Since the Q-values (of policy $\pi_{SR}$) need not be estimated, we do not include the reward layer in our architecture for learning the SF. Unlike other works \cite{machado2017laplacian,lakshminarayanan2016option}, Deep Successor options does not explicitly construct the graph and the formulation is hence naturally functional with neural networks. 

Once the SF is trained, a sample of the SF vectors is collected. Similar to the tabular case, the obtained vectors are clustered to produce SF cluster centers. These cluster centers represent various sub-goals. The intra-option policies are learnt using the reward function presented in Equation \ref{eq:fnrew}. In the equation, $g$ is the SF cluster centroid, and $\phi(s)$ is the intermediate feature representation. This formulation degenerates to the tabular setup when $\phi(s)$ is a one-hot vector. The reward function is based on an identical intuition where the options learn to navigate to landmark states (states with the highest value of $\psi_g \cdot \phi(s)$). As shown in Figure \ref{fig:net}, the options can be learnt using a separate head for each option (branch 3). Branch 2 remains frozen during the intra-option learning process, since it is responsible for determining the reward function for every option.

\begin{equation} \label{eq:fnrew}
    r_{\psi_g}(s_t, a_t, s_{t+1}) = \psi_g \cdot (\phi(s_{t+1}) - \phi(s_{t}))
\end{equation}

%% file: sections/exp.tex
\section{Experiments}

This section analyzes the answers to the following questions

\begin{figure}[bt]
    \centering
    
    \includegraphics[scale=0.17]{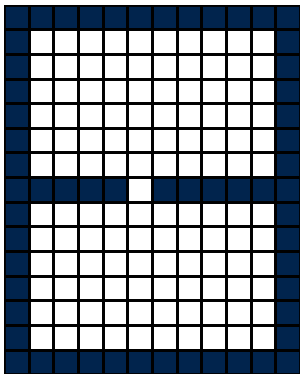}
    \includegraphics[scale=0.17]{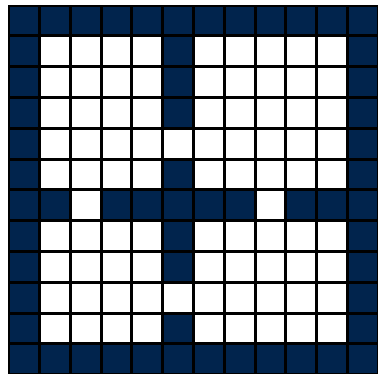} 
    \includegraphics[scale=0.10]{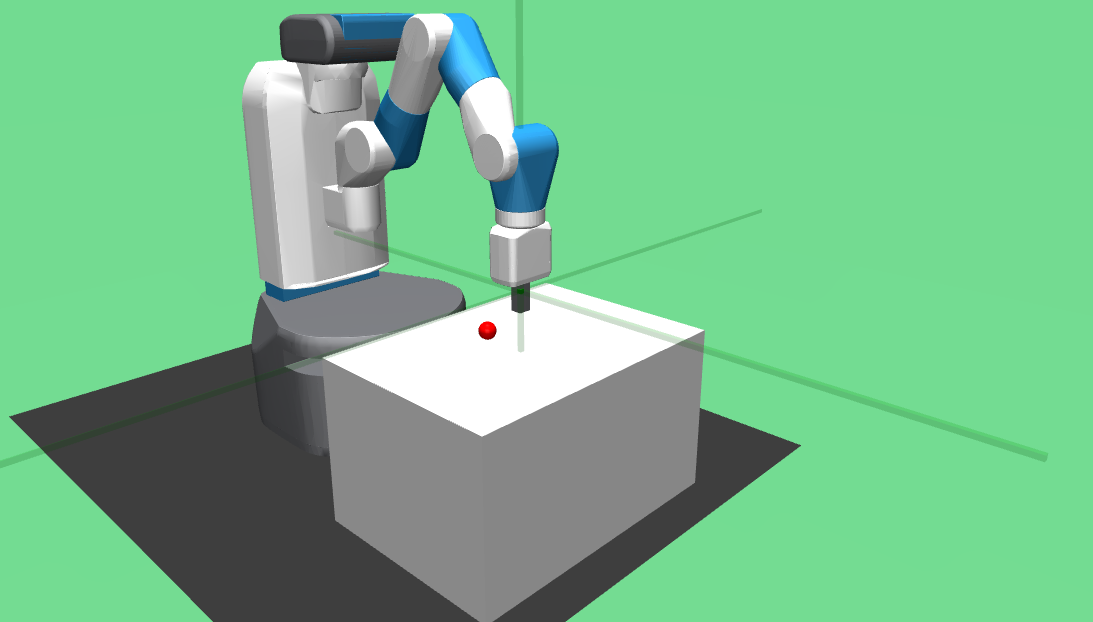}
    
    \vspace{0.1cm}
    
    \includegraphics[scale=0.17]{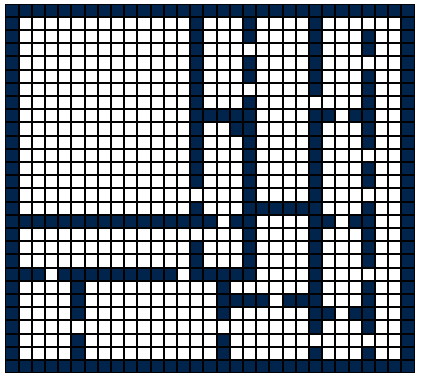}
    \includegraphics[scale=0.17]{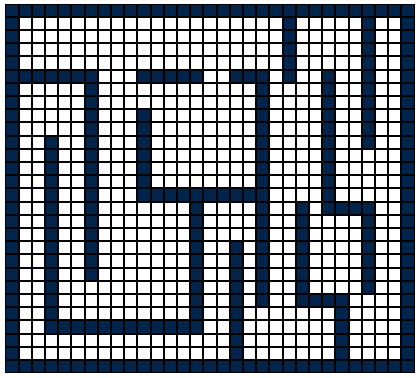}
    
    \caption{Grid-world environments and the Fetch-Reach task. The grid-worlds on the top (from left to right) are grid-1 and grid-2. The grid-worlds on the bottom (left to right) are grid-3 and grid-4.}
    \label{fig:envs}
\end{figure}

\begin{figure*}[ht]
    \centering
    \begin{subfigure}{.35\textwidth}
        \centering
        \includegraphics[scale=0.143]{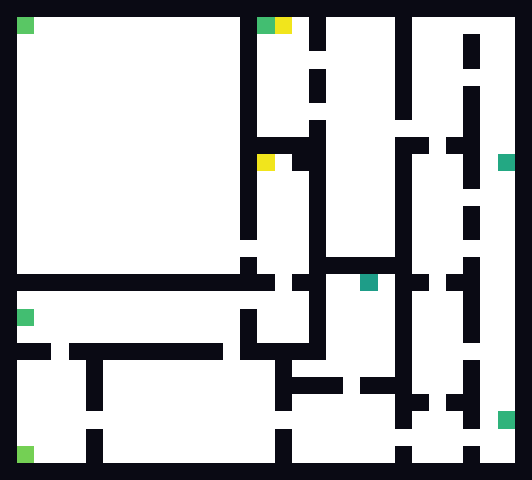} 
        \includegraphics[scale=0.143]{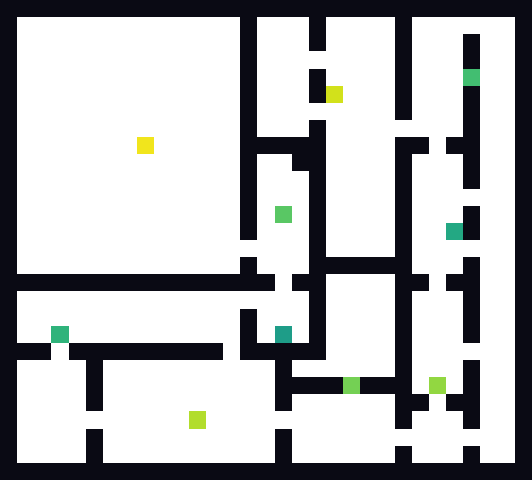}
        
        \includegraphics[scale=0.143]{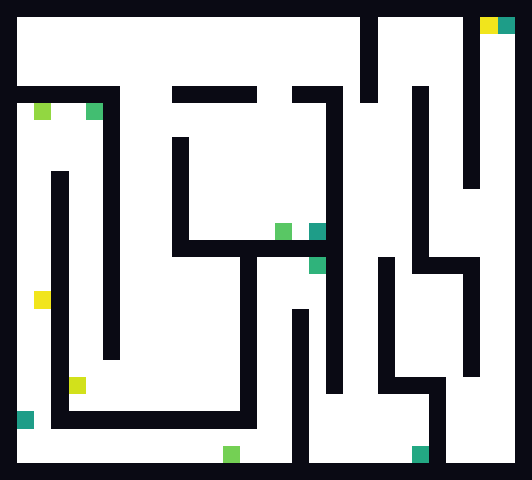} 
        \includegraphics[scale=0.143]{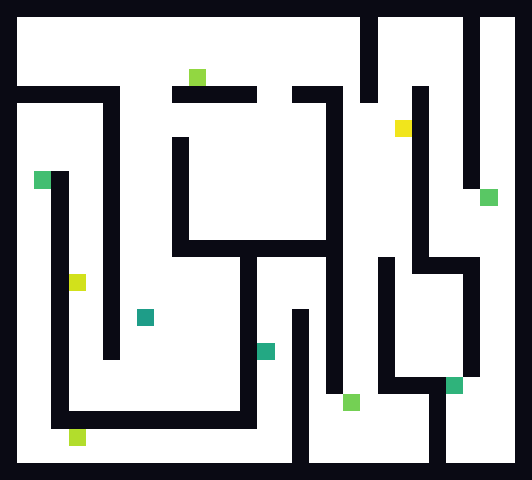}
        \caption{}
        \label{fig:subgoals}
    \end{subfigure}
    \begin{subfigure}{.18\textwidth}
        \centering
        \includegraphics[scale=0.19]{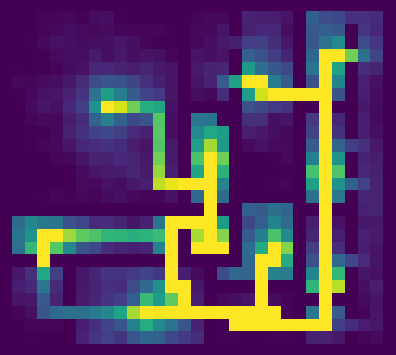}
        
        \includegraphics[scale=0.19]{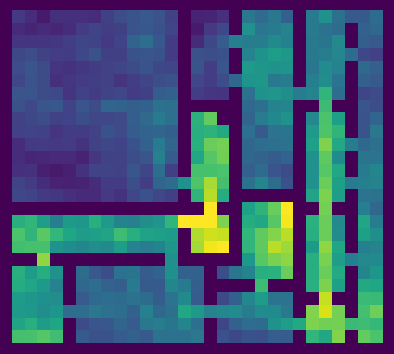}
        \caption{}
        \label{fig:heat}
    \end{subfigure}
    \begin{subfigure}{.40\textwidth}
        \centering
        \includegraphics[scale=0.16]{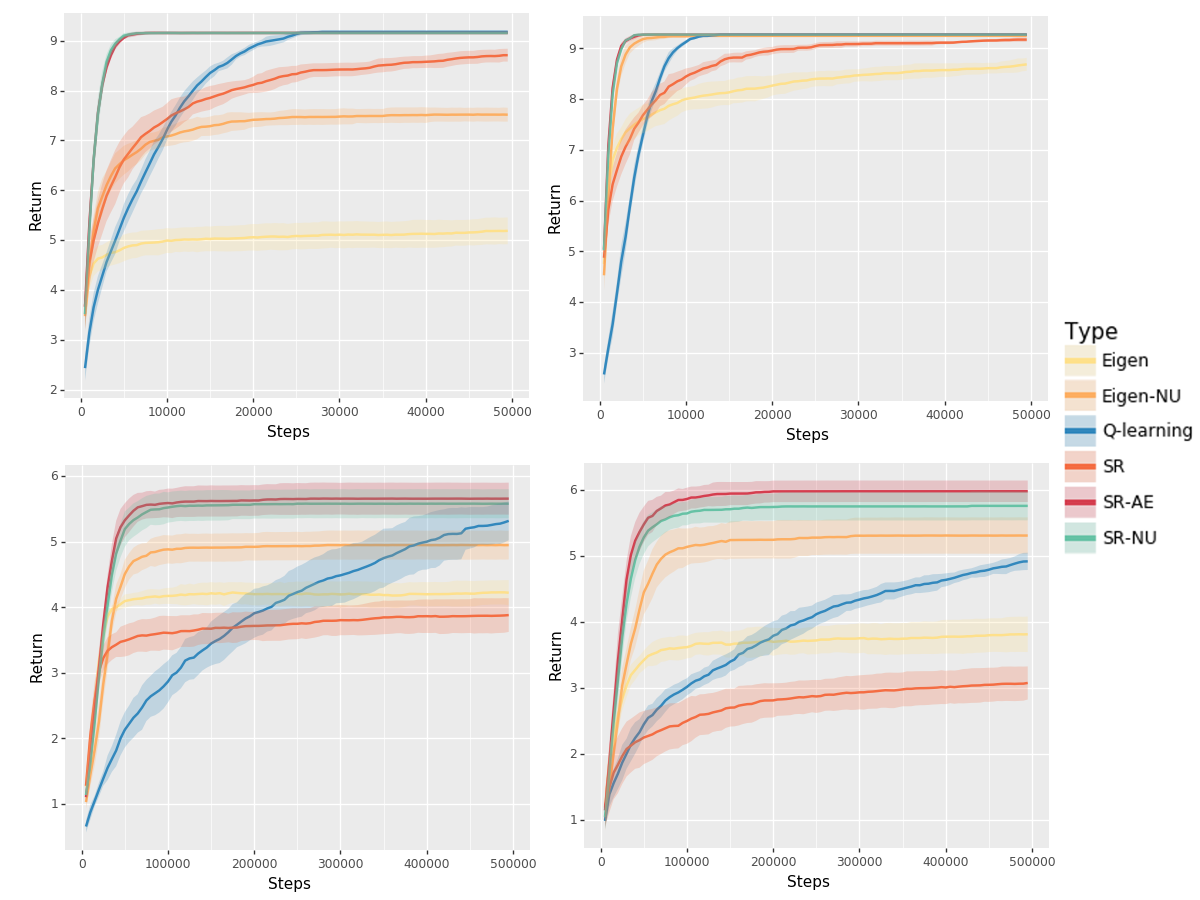}
        \caption{}
        \label{fig:eval}
    \end{subfigure}

    \caption{\textbf{(a)} Visualizing sub-goals for Eigen-options (left) and SR-options (right). The highlighted states are states where the options terminate (sub-goals). The different colours correspond to different options. Note that an option can terminate in multiple states and hence multiple states can have the same colour. \textbf{(b)} Heatmap of visitation counts. Image on top samples option:action in the ratio 1:1 and image on bottom samples options:actions in the ratio 1:500 \textbf{(c)} Performance plots for grid-1 (top-left), grid-2 (top-right), grid-3 (bottom-left) and grid-4 (bottom-right)}
\end{figure*}

% this is pushing conclusion to page 7, need to make it less verbose

 \begin{itemize}
     \item How different are the sub-goals discovered through SR-Options, from the ones discovered through other techniques such as Eigen-options? (Section \ref{sec:exp-subgoal})
     \item Why do we need a different exploration strategy when options are used? (Section \ref{sec:exp-expl})
     \item How do SR-Options fare empirically against other methods and baselines? (Section \ref{sec:exp-eval})
     \item How do SR-Options fare against Incremental SR-Options, in terms of discovered sub-goals in a finite horizon setting? (Section \ref{sec:exp-inc})
     \item And finally, how do SR-Options scale to handle continuous state and action spaces? (Section \ref{sec:exp-dso})
 \end{itemize}

\subsection{Tasks} \label{sec:tasks}

We consider 4 grid-world tasks, a finite horizon task, and the Fetch-Reach environment \cite{1802.09464}. There are 4 different grid-worlds (see Figure \ref{fig:envs}) with varying complexities. Each of them has 5 actions, the 5 being No-op, Left, Right, Up and Down. All transitions are completely deterministic. For the incremental setting, we consider grid-4 (from Figure \ref{fig:envs}) and limit the horizon to 100 steps.

For the first setup, we consider 500 random start and end states and evaluate on the same. The reward is +10 for reaching the goal and +0 otherwise and the discount factor $\gamma = 0.99$. For the second setup (finite horizon), we fix the start state to be the bottom-leftmost state and the goal to be the top-rightmost. The action space, reward, discount factor, and transition function are identical to that of the first setup. For the Fetch-Reach environment, we use the full state and action spaces of the task and use a $\gamma=0.99$

\subsection{Discovered sub-goals} \label{sec:exp-subgoal}

This section demonstrates the qualitative difference between SR-options and Eigen-options \cite{machado2017laplacian} through Figure \ref{fig:subgoals}. It is clear to see that the sub-goals are more diverse and spread out, for the case of SR-options. Furthermore, the discovered sub-goals are landmark states and situated in the middle of well-connected regions.

\subsection{Understanding Exploration with Options} \label{sec:exp-expl}

An SMDP optimal control framework typically uses options and actions together and explores using a uniform random policy between options and actions. eowever, options lead the agent to specific sub-goals unlike actions. As a result (as seen in Figure \ref{fig:heat}), the agent spends a majority of its time near these sub-goals. Hence, we propose two schemes for exploration when options are used. These two schemes are the Non-Uniform (NU) scheme and the Adaptive-Exploration (AE) scheme. In the NU scheme, options and actions are sampled in the ratio 1:$e$. Hence, the agent will navigate to sub-goals following which, a sequence of actions are used to explore the neighbourhood of that sub-goal. However, different neighbourhoods have different sizes. Since SR-options use a clustering step, the size of the cluster can be used to change the ratio at which options and actions are sampled. Hence, we propose the AE scheme where, after picking option $o_i$, options and actions are sampled in ratio  1 : $e \times \frac{\text{size of cluster} \ o_i}{\text{average size of cluster}}$. Hence, the sampling ratio is changed every time an option is picked and the agent makes use of the most recently used option to determine this ratio.

\subsection{Evaluating Successor Options} \label{sec:exp-eval}
This section highlights the quantitative differences between Eigen-options, Successor Options, and Q-learning. We have 6 different methods, namely Q-learning, SR-options, SR-options with NU scheme (SR-NU), SR-options with AE scheme (SR-AE), Eigen options and Eigen-options with NU scheme (Eigen-NU). We evaluate the first setup mentioned in Section \ref{sec:tasks}, for 5 different seeds. Each seed involves evaluating on 500 different random start and end-states. The grid-1 and grid-2 tasks are evaluated 100 times over 50,000 steps and grid-3 and grid-4 are evaluated 100 times over 500,000 steps. The number of options in grid-1, grid-2, grid-3, and grid-4 are (4, 5, 10, 10), but we have observed the performance to be robust to this hyper-parameter. The value of $e$ for the NU and AE schemes are (15, 15, 50, 50). We have observed that this parameter can be tuned further. The plots are presented in Figure \ref{fig:eval} and SR-AE has the best training-curve in all environments (with respect to area under curve and performance at t=0).

\subsection{Incremental Successor Options} \label{sec:exp-inc}

\begin{figure}[htb]
    \centering
    \includegraphics[scale=0.10]{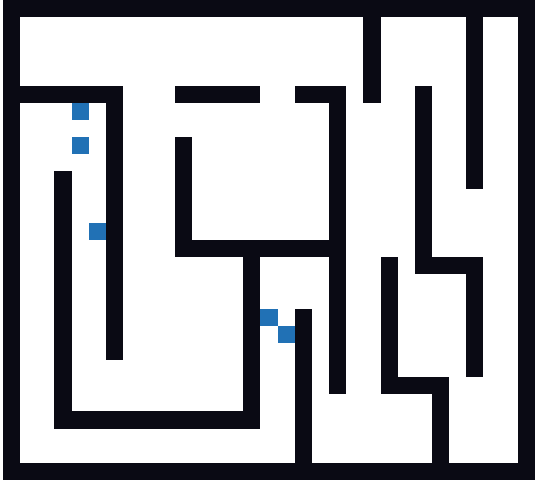}
    \includegraphics[scale=0.10]{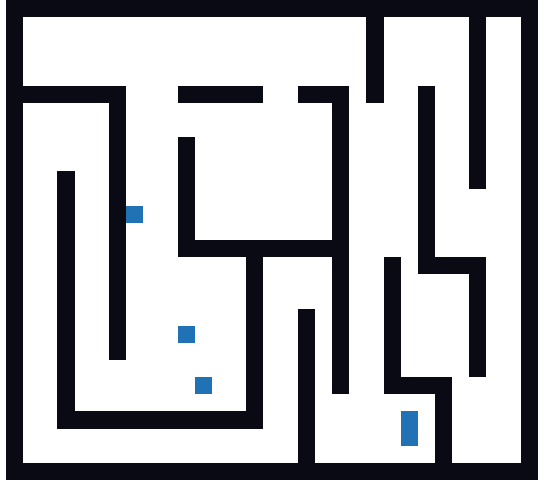}
    \includegraphics[scale=0.10]{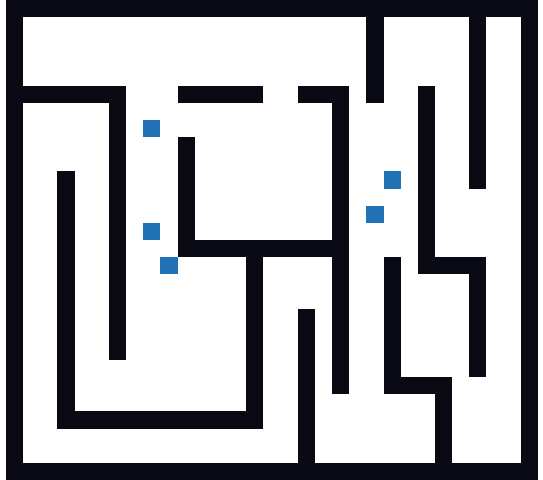}
    \includegraphics[scale=0.10]{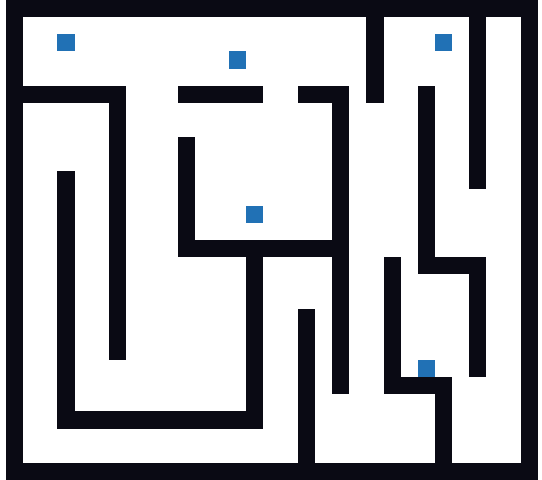}
    
    \includegraphics[scale=0.13]{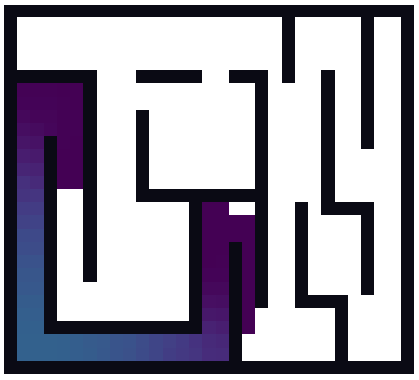}
    \includegraphics[scale=0.13]{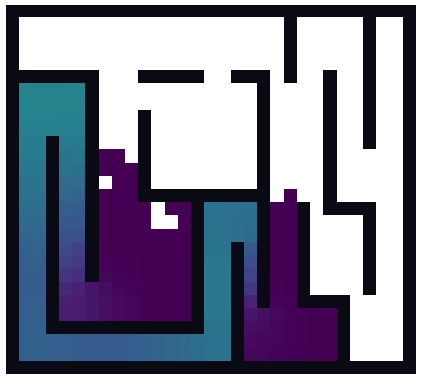}
    \includegraphics[scale=0.13]{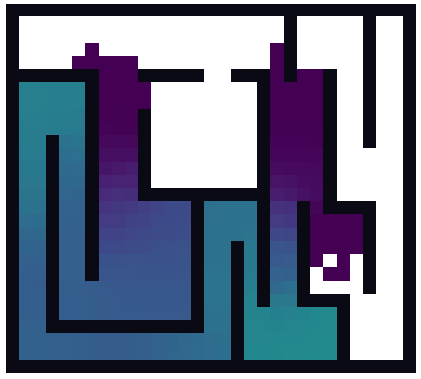}
    \includegraphics[scale=0.13]{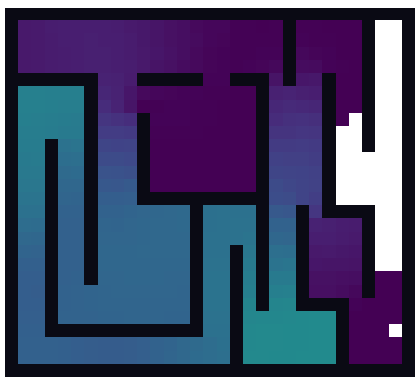}
    \caption{Bottom: The L1-norm of each state, for the first four iterations of the incremental SR algorithm, and Top: The discovered options (sub-goals) that augment the SR learning process}
    \label{fig:incsr}
\end{figure}

\begin{figure}[ht]
    \centering
    \includegraphics[scale=0.15]{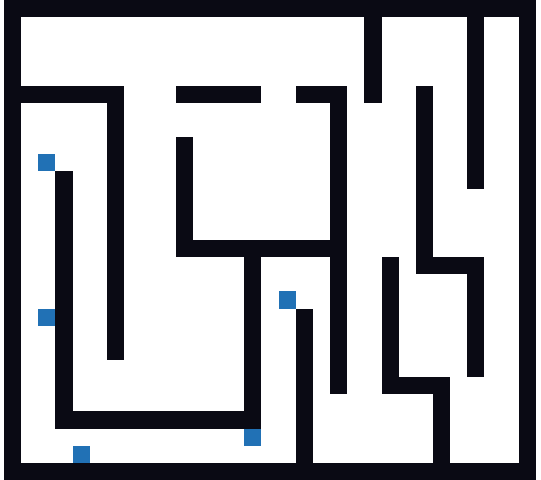} \hspace{0.15cm}
    \includegraphics[scale=0.15]{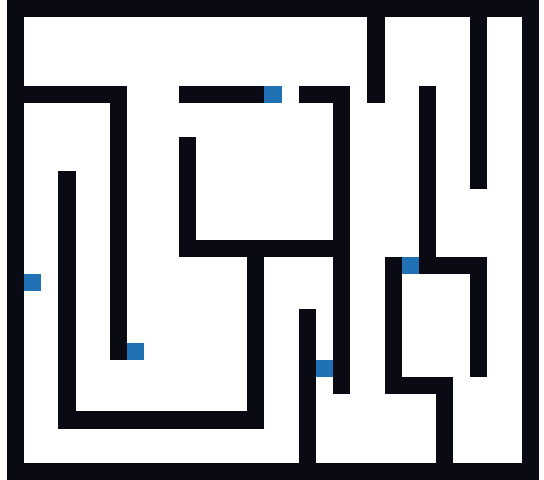}
    \caption{The final set of sub-goals from SR-options (left) and incremental SR-options (right), trained for same number of time-steps}
    \label{fig:incsr2}
\end{figure}

Incremental Successor Options is run using setup-2, described in section \ref{sec:tasks}, where the horizon is limited to 100 steps. Figure \ref{fig:incsr2} shows the nature of the discovered sub-goals when SR-options and incremental SR-options are used. Both algorithms are run for the same number of steps (intermediate option's learning time included). $SR_{min}$ is the 5\textsuperscript{th} percentile value and $SR_{max}$ is 40\textsuperscript{th} percentile value. Figure \ref{fig:incsr} plots the L1-norm of the SR-vectors for the first 4 iterations of training. We observe a clear increase in the explored state space in the given horizon, while discovering sub-goals that are well spread out.

\subsection{Understanding Deep Successor Options} \label{sec:exp-dso}
\begin{figure}[ht]
    \centering
    \includegraphics[scale=0.20]{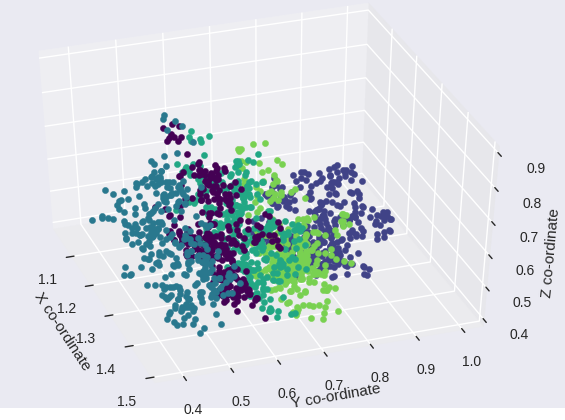}
    \caption{The end-effector positions of the different clusters}
    \label{fig:fetch}
\end{figure}

We look at the Fetch-Reach robotic control environment to look at the efficacy of Deep Successor Options. Figure \ref{fig:fetch} demonstrates that clustering over the Successor Representations naturally results in the segregation of the state space, based on the 3-dimensional co-ordinates. Moreover, we learn corresponding option policies (5 in total) using the intrinsic reward described in Equation \ref{eq:fnrew} and do so using the Proximal Policy Optimization (PPO) algorithm \cite{schulman2017proximal}. The option policies are observed to be diverse and have been visualized through a video \footnote{\url{https://www.dropbox.com/s/9284c190vlkimym/sroptions.mp4?dl=0}}

%% file: sections/related.tex
\section{Related Work}

\cite{moore1999multi} introduce airport hierarchies which assign different states as airports or landmarks with various levels defined on the basis of seniority. Each state is assigned to be a landmark only if it is reachable from a threshold number of states. The airport analogy is similar to the spread of clusters obtained from SR-options since each airport also represents a group of similar states. %Moreover, while learning iteratively, since we only use states with an adequately learnt SR, the candidate cluster centers are always able to leverage the SR as a dense reward and in turn learn successful option policies.

% The overwhelming majority of literature on option discovery employ techniques that revolve around discovering bottleneck states.
\cite{mcgovern2001automatic} describe a diverse density based solution that casts this problem as a multiple-instance learning task. The discovered solutions are bottlenecks since they are present in a larger fraction of positive bags. \cite{csimcsek2009skill} describe a betweenness centrality based approach which also naturally lead to bottleneck based options. Sub-goals based on relative novelty \cite{csimcsek2004using} identify states that could lead to vastly different states consequently which is closely tied to the notion of bottleneck states. Graph partitioning methods have also been employed to find options \cite{simsek2005identifying,lakshminarayanan2016option,menache2002q}. These methods design options that transition from one well-connected region to another. Since the sub-goals are the boundaries between two well-connected regions, these methods also typically identify bottlenecks as sub-goals. %In contrast, our work attempts to discover sub-goals that are representative of a well-connected region and are well separated in addition. 

Option Critic \cite{bacon2016optioncritic} is an end-to-end differentiable model that learns options on a single task. However, this method is forced to specialize for a single task and the learnt options are not easily transferable. Eigen-options \cite{machado2017laplacian} use the eigen-vectors of the Laplacian as rewards to learn intra-option policies. This method, however lacks a variety in sub-goals since ascending the different eigen-vectors often correspond to reaching the same sub-goal. The clustering step provides flexibility regarding the number of options required, which is absent in the case of eigen-options.  More recent work, \cite{machado2017eigenoption} attempts to use Successor Representations to obtain the eigen-vectors of the Laplacian. However, the obtained options are identical to the options obtained from Eigen-options (for reversible environments and under the uniform random policy) and hence our work significantly differs from this work. Successor Options clusters the SR vectors, while \cite{machado2017eigenoption} diagonalize the SR matrix to use the eigen-vectors of the same, in order to find the eigen-vectors of the graph Laplacian.

% If we have space, we can write more in conclusion/future work
% describe 
%  a) Inc options
%  b) Deep SR
%  c) Coming up with better termination condition (used probabilistic condition)
%  d) Using Compose-Net or other methods to compose different options together better

\section{Conclusion}
Successor Options is an option discovery framework that leverages Successor Representations to build options. Deep SR-options are formulated to work in the function approximation setting and the Incremental SR-options model attempts to address the finite horizon setting, where SRs cannot be constructed solely from primitive actions. 

As future work, we aim to use Deep Successor Options to achieve optimal control on high-dimensional sparse reward tasks. We believe that this is out of scope for this current work since high-dimensional spaces require a more robust termination condition and a reliable Successor Features network. This work assumes that the initiation set is the set of all states, which may not be an optimal choice. Another avenue for experimentation is to learn options, using a mixture of the pseudo-reward and an extrinsic reward. 
% Successor option